\documentclass[5p,authoryear]{elsarticle} 

\usepackage{framed,multirow}
\usepackage{amssymb}
\usepackage{latexsym}
\usepackage{xcolor}
\usepackage{lineno,hyperref}
\modulolinenumbers[5]
\usepackage{amsmath}
\usepackage{multirow}
\usepackage{color,graphicx}
\usepackage{nicematrix}
\usepackage{arydshln}
\usepackage{hyperref}
\usepackage{url}
\usepackage{footnote}
\usepackage{bm}
\usepackage{cite}
\usepackage{subfigure}
\usepackage{verbatim}
\usepackage{ulem}
\usepackage{graphicx}
\usepackage{pgfplots}
\usepackage{cancel}

\usepackage{booktabs}  
\usepackage{tabularx}

\usepackage{enumitem}

\journal{Medical Image Analysis}

\begin{document}

\begin{frontmatter}

\title{Personalized 3D Myocardial Infarct Geometry Reconstruction from Cine MRI for Cardiac Digital Twins}  

\author[label1]{Yilin Lyu} 
\author[label2,label3]{Mark YY Chan} 
\author[label2,label3]{Ching-Hui Sia} 
\author[label1]{Lei Li*} 
\ead[url]{lei.li@nus.edu.sg}

\address[label1]{Department of Biomedical Engineering, National University of Singapore, Singapore}
\address[label2]{Department of Medicine, National University of Singapore, Singapore}
\address[label3]{Department of Cardiology, National University Heart Centre Singapore, Singapore}

\begin{abstract}
Accurate 3D geometric characterization of myocardial infarction (MI) is essential for building cardiac digital twins (CDTs) to precisely simulate infarct-related electrophysiology. Late gadolinium enhancement magnetic resonance imaging (LGE MRI) is the clinical reference for locating MI, yet its reliance on contrast agents restricts use in renally impaired patients and limits longitudinal follow-ups. As an alternative, contrast-free cine MRI visualizes abnormal ventricular wall motion, which is highly indicative of the infarcted area. In this study, we propose a novel explicit geometry-motion embedded model to fully automatically reconstruct personalized, simulation-ready 3D MI geometries directly from multi-view cine MRIs. Specifically, we construct a 4D (3D + t) biventricular mesh to explicitly extract and decouple geometry-aware and motion-aware features. We further design a dual-branch module for adaptive geometry-motion fusion to capture spatiotemporal dependencies for mapping infarcted region. Furthermore, we introduce multi-scale supervision utilizing an AHA-17 segment-guided cross-attention mechanism to steer the prediction, ensuring biophysically consistent reconstruction. Experimental results on 225 cine MRIs demonstrated that the proposed 3D MI reconstruction achieved high performance with an average Dice score of 0.678 $\pm$ 0.011. In the downstream in-silico electrophysiological simulation evaluations, the results were highly consistent with the LGE-derived ground truth, highlighting the great potential of the proposed model for contrast-free scar characterization and seamless integration into CDT modeling. The code will be released publicly upon acceptance of the manuscript for publication.

\end{abstract}

\begin{keyword}
Myocardial Infarction \sep Cine MRI \sep 3D Infarct Reconstruction \sep Contrast Free \sep Electrophysiological Simulation \sep Cardiac Digital Twins  
\end{keyword}

\end{frontmatter}

\section{Introduction}

Myocardial infarction (MI) remains a major cause of sudden cardiac death worldwide \citep{journal/Lancet/reed2017}.
The size, location, and transmurality of infarcted region, also referred to as myocardial scars, are key determinants of patient-specific arrhythmic risk in post-MI patients \citep{journal/NC/arevalo2016,conf/MICCAI/sheng2025}.
Yet, current clinical risk stratification relies on left ventricular (LV) ejection fraction, a global metric that fails to capture these scar-related features \citep{journal/FCM/vancheri2024,journal/EP/silvola2025}. 
This mismatch leads to inaccurate evaluation of SCD risk and patient selection for primary prevention implantable cardioverter-defibrillator treatment \citep{journal/Europace/kolk2023,journal/JACC/wallace2025}. 
Cardiac digital twins (CDTs) offer a promising solution by enabling patient-specific computational modeling of post-MI electrophysiology \citep{journal/NC/arevalo2016,journal/NEJM/chrispin2026}.
A key step of CDT is high-fidelity anatomical modeling, which serves as the prerequisite for accurate functional simulations \citep{journal/MedIA/gillette2021}. 
Besides the geometric reconstruction of the heart itself, the accurate 3D reconstruction of the infarct regions is also necessary for electrophysiological (EP) modeling of post-MI \citep{journal/JMCC/codreanu2008,journal/EP/wang2021,journal/TMI/li2024}.
This is because infarcted regions alter local electrical conductivity and pathways of wave propagation \citep{journal/CR/rutherford2012}.

Late gadolinium enhancement magnetic resonance imaging (LGE MRI) is the non-invasive clinical reference for visualizing myocardial scar \citep{journal/TMI/ukwatta2015}, yet its use in CDT construction entails a fragmented and labor-intensive workflow \citep{journal/JACC-CI/fahmy2018}.
This pipeline typically involves reconstructing biventricular anatomy, segmenting scar from 2D LGE slices, and interpolating the segmented regions onto 3D anatomical geometries \citep{journal/NC/arevalo2016,journal/CAE/waight2025}. 
A fundamental limitation is that LGE MRI typically captures only sparse, intersecting 2D planes, yielding coarse scar representations that may fail to capture detailed pathological topology and local conduction heterogeneity \citep{journal/NC/arevalo2016,journal/JMRI/toupin2022}. 
The subsequent interpolation between these sparse slices assumes smooth and continuous scar topology, which rarely holds for the complex, irregular, and often discontinuous morphology of real infarct tissue \citep{journal/TMI/ukwatta2015}.
Although 3D high-resolution LGE MRI can provide finer scar detail \citep{journal/CAE/waight2025,journal/NEJM/chrispin2026}, it remains rare in clinical practice due to high costs and lengthy acquisition times \citep{journal/JMRI/basha2017}.
Furthermore, gadolinium contrast poses contraindications for certain patient groups (e.g., those with renal insufficiency) \citep{journal/MRMPBM/parillo2024}, thereby limiting the scalability and clinical integration of CDT modeling.
In contrast, cine MRI is routinely available with rich 2D multi-view acquisitions (e.g., long-axis and short-axis stacks), and encodes not only anatomical shape but also myocardial motion throughout the cardiac cycle \citep{journal/MedIA/li2023}. 
Note that the presence of scars is closely associated with characteristic abnormalities in the regional motion pattern \citep{journal/CAR/Feldmann2019,journal/CHFR/thune2006,journal/IHJ/cerisano2001}.
Therefore, several studies employed the abnormal cardiac motion information captured by cine imaging to infer infarcted regions \citep{journal/MedIA/xu2018,journal/Radiology/zhang2019,conf/ISBI/yang2025}.
These methods, however, typically produce only 2D image-level predictions, which are insufficient for accurate 3D EP simulation of MI.

In this study, we develop a novel contrast-free 3D infarct geometry reconstruction model that leverages spatiotemporal features extracted from routine multi-view cine MRI, namely GeoMo-Net. 
Unlike existing approaches that either synthesize LGE-like images or detect infarcts from 2D motion abnormalities, our framework explicitly integrates cardiac morphology and motion to establish a direct, learnable mapping between abnormal myocardial motion and 3D infarcted regions. 
Specifically, we formulate this task as node-level infarction localization on the cardiac mesh and explicitly integrate geometry-aware features that encode anatomical substrate and motion-aware features that capture dynamic dysfunction.
This geometry-motion decoupled representation preserves the distinct physical meanings of structure and function while providing an interpretable basis for learning the relationship between abnormal myocardial motion and underlying scar distributions.
We further incorporate multi-scale anatomical supervision to improve both local boundary fidelity and segment-level consistency.
To the best of our knowledge, this is the first study that directly reconstructs simulation-ready, scar-embedded 3D cardiac models from routine cine MRI for post-MI EP modeling.
The main contributions of this work are summarized as follows:
\begin{enumerate}[label=\roman*.]
\item We propose a fully automatic 3D MI reconstruction framework that enables direct construction of simulation-ready anatomical models from routinely available cine MRI, eliminating the need for LGE MRI during inference.

\item We design a geometry-motion decoupled representation that integrates myocardial morphology, displacement, velocity, and strain, enabling direct mapping from abnormal motion patterns to 3D infarct locations.

\item We formulate scar-embedded 3D cardiac reconstruction as a spatio-temporal learning problem, where the mapping from motion patterns to scar distributions is regularized by multi-scale anatomical priors.

\item We demonstrate that the reconstructed 3D infarct geometry can provide comparable EP simulation results to those obtained from LGE-derived ground truth.
\end{enumerate}

This study is a systematic extension of our previous workshop work \citep{conf/DT4H/lyu2025}. First, we propose a more complete geometry-motion learning framework with explicit geometry- and motion-aware feature decoupling, spatio-temporal modeling, and AHA-17 segment-guided multi-scale supervision. Second, we provide more comprehensive quantitative evaluation, including comparisons with multiple representative baselines, ablation studies, scar localization and burden estimation, and correlation analysis between mesh reconstruction quality and scar prediction accuracy. Third, we further validate the downstream functional relevance of the reconstructed scars using in-silico EP simulation. Finally, we include external validation on the public dataset to assess cross-domain generalizability. 

\section{Related Work}

\subsection{Contrast-Free Inference of Myocardial Infarction}

Inferring myocardial scar without gadolinium-based contrast agents has gained increasing attention \citep{journal/Cir/zhang2022,journal/TMI/li2024}. 
Existing studies have investigated three major non-contrast modalities: cine MRI \citep{journal/FiCM/cicek2024,conf/ISBI/yang2025}, echocardiography (echo) \citep{journal/FiCM/nguyen2023,journal/BSPC/degerli2024}, and electrocardiogram (ECG) \citep{journal/IS/acharya2017,journal/SR/boribalburephan2024}, as well as their combinations \citep{journal/TMI/li2024}. 
These approaches target varying output representations of scars, including 2D segment-level classification, pixel-level segmentation, and 3D modeling.
Echo and ECG offer complementary physiological or electrical insights, but they are generally limited to 17-AHA segment-level scar classification. 
For echo, the limited spatial resolution and image quality often preclude precise pixel-level delineation of scar boundaries \citep{journal/BSPC/degerli2024}. 
For ECG, the inverse problem of reconstructing cardiac sources from body-surface potentials is inherently ill-posed, making pixel-level localization of scar regions challenging \citep{journal/SR/boribalburephan2024,journal/TMI/li2024}. 
Consequently, both modalities are typically used for coarser identification of which AHA segments are likely infarcted, rather than for fine-grained scar delineation.

Cine MRI has therefore become an important non-contrast modality for finer-grained infarct inference, benefiting from its routine availability and dynamic depiction of myocardial motion \citep{journal/EHJIMP/elshibly2025}.
Related cine MRI-based scar detection methods fall into two categories. 
The first category synthesizes LGE-like images from cine sequences \citep{journal/Cir/zhang2022,journal/CCI/qi2024,journal/MedIA/qi2025,journal/CCI/qi2026}.
Although image synthesis provides an intuitive surrogate for contrast-enhanced scar visualization, its performance may be sensitive to domain shifts across scanners, acquisition protocols, and patient cohorts.
The second category directly infers infarct-related regions based on cardiac motion or its implicit representation for 2D scar segmentation \citep{journal/Radiology/zhang2019,journal/FiCM/abdulkareem2022, journal/Diagnostics/ali2026}. 
These outputs often lack explicit 3D biventricular geometry, lesion topology across slices, and mesh compatibility, all of which are essential for computational EP modeling.
Recent point cloud-based methods have explored 3D scar geometric modeling from 2D cine MRI \citep{journal/ASC/liu2023, journal/TMI/li2024}. 
While these methods move beyond 2D prediction, point clouds represent scars as unstructured spatial samples and lack explicit surface connectivity, volumetric elements, and temporally consistent mesh topology.

\begin{figure*}[t]
\centering
\includegraphics[width=0.94\textwidth]{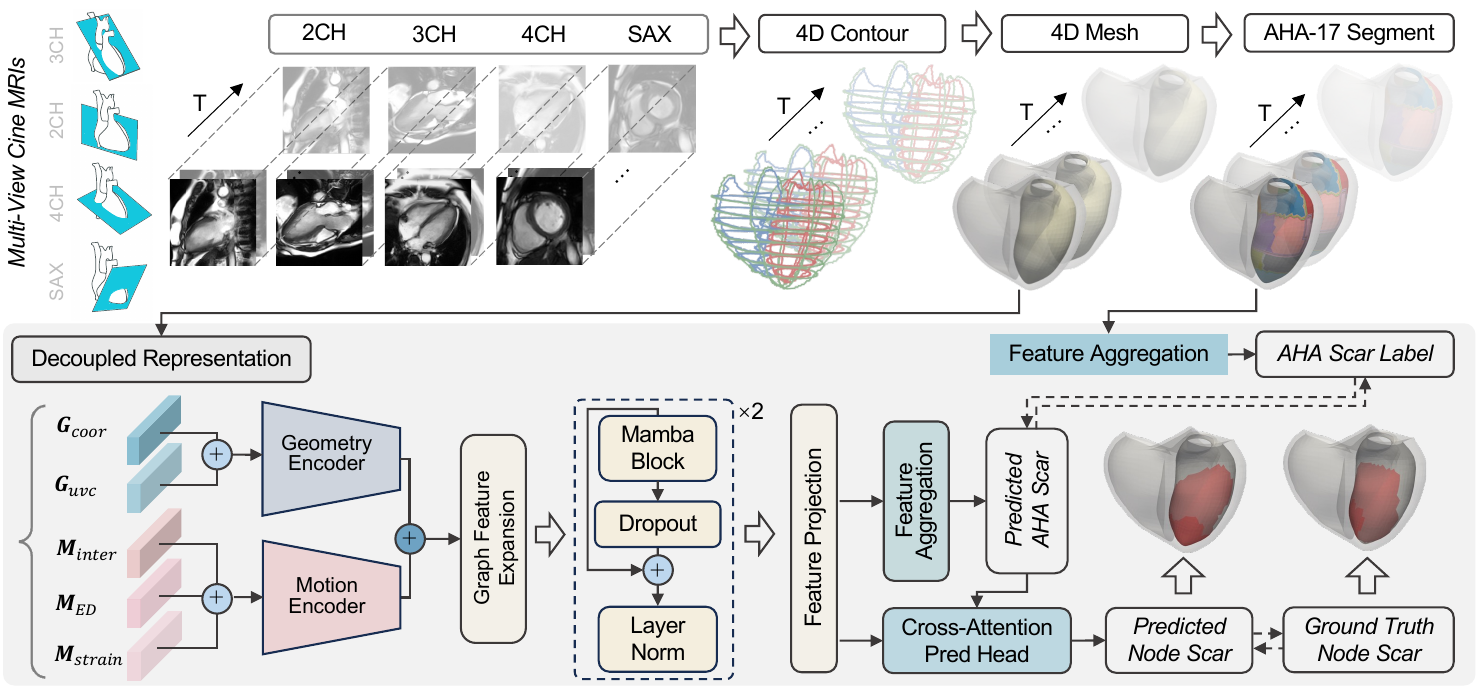}\\[-2ex]
\caption{Pipeline of the proposed 3D myocardial infarct geometry reconstruction framework from multi-view cine MRIs, including 2CH, 3CH, 4CH (2-, 3-, and 4-chamber long-axis views) and SAX (short-axis views). }
\label{fig:method:framework}
\end{figure*}

\subsection{Cardiac Electrophysiology Simulation of Myocardial Infarction for Digital Twinning}

Digital twins provide a patient-specific virtual-heart platform for post-MI risk stratification and treatment planning by integrating ventricular anatomy, infarct remodeling, and EP function. 
For example, virtual-heart arrhythmia risk predictors have been used to estimate sudden cardiac death risk by performing multi-site virtual pacing around remodeled infarct tissue, outperforming conventional clinical metrics in predicting future arrhythmic events \citep{journal/NC/arevalo2016}. 
Beyond risk prediction, MI digital twins have also been applied to ventricular tachycardia ablation planning.
On the digital heart, virtual pacing was used to identify inducible circuits and optimize ablation targets in a risk-free simulation environment \citep{journal/Nature_BME/prakosa2018,journal/Europace/bhagirath2023}. 
These studies demonstrate the clinical potential of EP simulation for evaluating infarct-related substrates and guiding individualized therapy. 

Existing pipelines typically delineate infarct regions from LGE MRI and register the resulting scar maps onto patient-specific ventricular meshes for simulation \citep{journal/NC/arevalo2016,journal/Europace/bhagirath2023,journal/Circ/waight2025}. 
While such approaches have demonstrated value for arrhythmic substrate characterization, they often depend on multi-stage processing, including anatomical segmentation, scar delineation, image registration, mesh generation, topology correction, and tissue property assignment \citep{journal/JCTR/vandenbroek2019,journal/PR/trayanova2024}. 
Recent deep learning-based methods have improved cardiac segmentation, surface reconstruction, and topology-aware mesh generation \citep{journal/TMI/kong2022,journal/TMI/pak2023,journal/CBM/ahmet2025}, and emerging work has begun to integrate scars into personalized simulation models of post-MI \citep{journal/TMI/li2024}. 
However, current studies still treat anatomical reconstruction and scar modeling as separate steps or rely on contrast-enhanced imaging. 
A direct cine MRI-based framework that produces scar-embedded and simulation-ready biventricular models therefore remains highly desirable for scalable MI digital twinning.

\section{Methodology}

Figure \ref{fig:method:framework} provides the overall pipeline of the proposed framework that consists of three main components.
First, a topology-consistent 4D biventricular mesh is generated from cine MRI, and LGE MRI is spatially aligned with the cine-derived geometry to provide scar supervision during training (Sec. \ref{method:mapping}).
Second, geometry- and motion-aware features are explicitly extracted from the dynamic mesh to represent both anatomical substrate and functional abnormality (Sec. \ref{method:decoupling}).
Third, spatio-temporal modeling is employed to produce dense scar predictions under multi-scale supervision (Sec. \ref{method:recon}).

\subsection{Topology-Consistent 4D Mesh Reconstruction and Scar Projection} \label{method:mapping}

To establish a simulation-compatible anatomical domain, we first reconstruct a 4D surface mesh from cine MRI via an open-source biventricular modelling pipeline \citep{conf/FIMH/bivme2025}.
Specifically, a pretrained nnU-Net automatically segments biventricular structures and extracts contours, which are then mapped to a unified world coordinate system. 
Biventricular meshes for each frame are reconstructed via iterative diffeomorphic registration, yielding a node-correspondent 4D surface mesh representation that is topologically consistent across the cardiac cycle.
This node correspondence allows anatomical labels defined at the end-diastolic (ED) frame to be associated with the same node indices throughout the full cine sequence.
To generate 3D infarct supervision labels, we project segmented LGE MRI scars onto the 3D biventricular surface mesh reconstructed from cine MRI. 
A multivariate mixture model-based registration framework \citep{journal/TPAMI/zhuang2018} is used to correct misalignment between the LGE and cine MRIs caused by differences in resolution, field of view, and respiratory motion.
Once LGE MRI is spatially aligned with cine MRI, the LGE-derived scar annotations can be transformed into the cine coordinate system and subsequently associated with the ED biventricular surface mesh. 
Nonetheless, the transformed infarct regions remain spatially sparse and discontinuous due to the large inter-slice gaps of 2D LGE MRI acquisitions.
Therefore, an additional scar-to-surface mapping procedure is required to convert the registered scar annotations into dense node-wise labels on the LV endocardial surface.

We propose a spatial mapping algorithm to project 2D sparse scar masks onto the 3D biventricular surface at the ED frame, both represented in a unified space. 
To approximate the anatomical continuity of myocardial scars, the registered sparse masks are first resampled into an isotropic 3D volume using trilinear interpolation.
We then convert the resampled masks into a point cloud, followed by 3D spatial interpolation using anisotropic diffusion with a 3D Gaussian kernel. 
This ensures plausible continuous scar geometry between adjacent slices while smoothing scar boundaries. 
The reconstructed volumetric scar representation is then projected onto the LV endocardial surface. 
Because a conventional spherical neighborhood search is prone to crosstalk from adjacent or contralateral wall segments and sensitivity to registration errors, we design a normal-aligned cylindrical search strategy. 
For each endocardial node, a one-sided cylindrical search region is defined with the node as the base center and the inward normal vector as the central axis. 
The displacement of each candidate scar point is decomposed into an axial projection distance along the normal and a radial perpendicular distance, and the point is included only if both distances fall below preset thresholds. 
Furthermore, to provide auxiliary supervisory signals at the regional scale, we construct a segment-level scar burden prior using the AHA 17-segment model \citep{journal/Circ/aha2002}. 
Each LV endocardial node is assigned circumferential ($\xi_{\text{circ}} \in [0,1]$) and longitudinal ($\xi_{\text{long}} \in [0,1]$) coordinates in the universal ventricular coordinates (UVC) system \citep{journal/MedIA/bayer2018}, which encode its rotational and apex-to-base positions, respectively. 
Following the AHA partitioning scheme, each node is allocated to a specific segment. 
The scar burden for each segment is derived by averaging the binary scar labels of its constituent nodes, yielding an anatomically interpretable coarse-grained prior for multi-scale supervision.

\subsection{Explicit Geometry-Motion Decoupled Representation} \label{method:decoupling}

We formulate scar prediction as a node-level binary classification problem on topology-consistent 4D surface meshes. 
Let $\mathcal{G} = (\mathcal{V}, \mathcal{E})$ denote the 3D heart in terms of mesh with $N = |\mathcal{V}|$ nodes and connectivity $\mathcal{E}$ extracted from triangular faces. 
Each node $v_i \in \mathcal{V}$ is continuously tracked across $T$ cardiac phase frames, forming a 4D spatiotemporal sequence $\mathbf{X} \in \mathbb{R}^{T \times N \times 3}$. 
Given the dynamic mesh sequence, the objective is to predict a continuous scar probability for each node $i$, which is subsequently converted into a binary label $\hat{y}_i \in \{0, 1\}$ to indicate the presence or absence of infarcted tissue. 
Infarcted regions may exhibit structural abnormalities such as wall thinning, as well as functional abnormalities including reduced contraction and paradoxical motion \citep{journal/Circ/corya1977,journal/Circ/sutton2000,journal/Circ/ross1986}. 
Simply concatenating all available features into a single representation may therefore obscure the distinct physical meanings of anatomical geometry and myocardial dynamics. 
To preserve this distinction, we explicitly decouple the input representation into a geometry-aware branch and a motion-aware branch.

The geometry-aware branch encodes the anatomical substrate of the dynamic mesh.
The primary geometry representation is the coordinate of each node, defined as $\mathbf{G}_{\text{coor}} = \{\mathbf{x}_i^{t}\}_{t=0, i=1}^{T-1, N} \in \mathbb{R}^{T \times N \times 3}$, where $\mathbf{x}_i^{t}$ denotes the 3D coordinates of node $i$ at cardiac phase $t$. 
In addition, we incorporate UVC-based spatial coordinates as anatomical priors: $\mathbf{G}_{\text{uvc}} = \{(\xi_{\text{circ},i}, \xi_{\text{long},i})\}_{i=1}^{N} \in \mathbb{R}^{N \times 2}$. 
These provide a normalized representation of circumferential and longitudinal position, enabling the model to distinguish anatomically distinct regions that may share similar local geometry. 
The mesh connectivity is extracted from the triangular faces and remains fixed across all cardiac phases, providing the graph structure to aggregate local anatomical context. 
The motion-aware branch characterizes dynamic myocardial dysfunction related to infarction using three complementary motion descriptors. 
First, inter-frame displacement captures local instantaneous motion between adjacent cardiac phases $\mathbf{M}_{\text{inter}}$. 
It is computed via first-order finite difference of node coordinates along the temporal axis:
\begin{equation}
\mathbf{M}_{\text{inter}}^{(t,i)} = 
\begin{cases}
\mathbf{x}_i^{t} - \mathbf{x}_i^{t-1}, & t \geq 1 \\
\mathbf{0}, & t = 0
\end{cases}
\end{equation}
where $\mathbf{M}_{\text{inter}} \in \mathbb{R}^{T \times N \times 3}$ and the initial frame is padded with zeros. 
Second, reference-frame displacement, denoted as $\mathbf{M}_{\text{ED}} \in \mathbb{R}^{T \times N \times 3}$, describes accumulated deformation over the cardiac cycle. 
It is defined as the spatial offset of each node at frame $t$ relative to the ED reference frame at $t=0$ (i.e., $\mathbf{x}_i^{t} - \mathbf{x}_i^{0}$). 
Unlike inter-frame displacement, this motion descriptor provides a macroscopic representation of regional contraction or abnormal motion throughout the entire cardiac cycle.
Third, to further capture local deformation patterns, we compute edge-wise surface strain on the mesh. 
For each edge $(i,j) \in \mathcal{E}$, the normalized length change at frame $t$ is defined as:
\begin{equation}
\varepsilon_{ij}^{t} = \frac{\|\mathbf{x}_i^{t} - \mathbf{x}_j^{t}\|_2 - \|\mathbf{x}_i^{0} - \mathbf{x}_j^{0}\|_2}{\|\mathbf{x}_i^{0} - \mathbf{x}_j^{0}\|_2 + \epsilon}
\end{equation}
where $\epsilon$ is a small constant added for numerical stability. 
The edge-wise strain values are then aggregated to their adjacent nodes via averaging to obtain the nodal strain descriptor $\mathbf{M}_{\text{strain}} \in \mathbb{R}^{T \times N}$. 
This emphasizes local deformation inconsistency and may reflect subtle functional impairment around scarred myocardium. 
Together, the geometry-aware representation set $\{\mathbf{G}_{\text{coor}}, \mathbf{G}_{\text{uvc}}\}$ describes the anatomical substrate, whereas the motion-aware representation set $\{\mathbf{M}_{\text{inter}}, \mathbf{M}_{\text{ED}}, \mathbf{M}_{\text{strain}}\}$ describes motion abnormality. 
This explicit separation provides a physically interpretable representation for subsequent scar reconstruction \citep{conf/ICCV/yuan2023}.

\subsection{Spatio-Temporal Modeling for 3D Infarct Reconstruction} \label{method:recon}

The extracted geometry and motion representation are fed into a dual-branch graph neural network designed for node-level infarct prediction. 
As illustrated in Fig.~\ref{fig:method:framework}, it consists of three main components: a dual-branch encoder with adaptive fusion, a spatial-temporal modeling module, and a multi-scale prediction head.
The geometry branch receives the concatenated geometry features, i.e., cardiac mesh coordinates $\mathbf{G}_{\text{coor}}$ and spatial priors $\mathbf{G}_{\text{uvc}}$. 
A graph convolutional network (GCN) layer maps this input to a geometric embedding. 
In parallel, the motion branch processes the motion features $\mathbf{M}_{\text{inter}}$, $\mathbf{M}_{\text{ED}}$, and $\mathbf{M}_{\text{strain}}$ through separate GCN layers, producing a cardiac motion embedding. 
Both branches use the shared mesh connectivity $\mathcal{E}$ to aggregate local neighborhood information on the surface mesh.
To balance the contribution of structural and functional information, we introduce an adaptive fusion mechanism with learnable weights normalized via Softmax.
The fused representation is then passed through a GCN-based expansion module to aggregate higher-order local context from neighboring nodes.
To model temporal dynamics across the cardiac cycle, the expanded node features are processed as node-wise temporal sequences.
They are fed into two stacked Mamba \citep{arxiv/gu2023} modules with shared parameters across nodes, allowing the model to capture full-cycle temporal dependencies.
The temporally modeled features are then compressed via linear projection into a latent tensor, followed by temporal average pooling to obtain a static full-cycle representation. 

We employ a multi-scale supervision strategy to combine dense node-level prediction with regional anatomical regularization.
For coarse-scale supervision, we use the AHA 17 segment-level scar burden derived from the projected node-wise labels. 
Node features are aggregated within each segment via mask-weighted average pooling, yielding segment-level features. 
A lightweight multi-layer perceptron (MLP) then predicts the segment-level scar probability vector. 
To propagate this regional context to individual nodes, we introduce a unidirectional cross-attention module, where node features attend to their corresponding segment representation. 
This allows each node to incorporate anatomical context from its corresponding AHA segment while preserving node-level spatial specificity. 
The refined node features are then used for dense-scale prediction.
A graph-level global max-pooled feature is first computed from all node features and broadcast back to each node.
The broadcast global feature is concatenated with the refined node feature along the channel dimension and passed through an MLP to output node-wise scar probabilities.
The overall loss function combines dense node-level supervision with coarse anatomical regularization. 
Due to severe class imbalance (scar regions typically occupy a small fraction of the myocardial surface), the dense loss integrates Dice loss and Focal loss. 
The coarse loss uses mean squared error (MSE) to supervise the AHA-level scar probabilities. 
The total objective is:
\begin{equation}
\mathcal{L}_{\text{total}} = \mathcal{L}_{\text{Dice}} + \lambda_{\text{focal}} \mathcal{L}_{\text{focal}} + \lambda_{\text{coarse}} \mathcal{L}_{\text{MSE}}
\end{equation}
where $\lambda_{\text{focal}}$ and $\lambda_{\text{coarse}}$ are balancing parameters.

\section{Experiments and Results}

\begin{table*}[htbp]
\centering
\caption{Summary of the quantitative evaluation results of 3D infarct reconstruction using different methods. G-Dice: generalized Dice. ASSD: average symmetric surface distance. } 
\label{exp:tb:comparison}
{\scriptsize
\setlength{\tabcolsep}{4pt}
\begin{tabular}{l| c c c c c c c}
\hline
Method & Dice $\uparrow$ & G-Dice $\uparrow$ & Sensitivity $\uparrow$ & Specificity $\uparrow$ & Precision $\uparrow$ & Accuracy $\uparrow$ & ASSD (mm) $\downarrow$ \\
\hline
GAT \citep{conf/ICLR/velickovic2018} & 0.612 $\pm$ 0.057 & 0.688 $\pm$ 0.028 & 0.702 $\pm$ 0.103 & 0.803 $\pm$ 0.053 & 0.591 $\pm$ 0.040 & 0.775 $\pm$ 0.021 & 8.416 $\pm$ 3.232 \\
PointNet++ \citep{conf/NIPS/qi2017} & 0.609 $\pm$ 0.032 & 0.671 $\pm$ 0.027 & 0.716 $\pm$ 0.059 & 0.770 $\pm$ 0.048 & 0.554 $\pm$ 0.037 & 0.754 $\pm$ 0.027 & 8.199 $\pm$ 3.445 \\
ST-GCN \citep{conf/AAAI/yan2018} & 0.626 $\pm$ 0.011 & 0.691 $\pm$ 0.012 & 0.747 $\pm$ 0.027 & 0.777 $\pm$ 0.026 & 0.571 $\pm$ 0.022 & 0.769 $\pm$ 0.013 & 8.953 $\pm$ 3.671 \\
C2I-Net \citep{conf/DT4H/lyu2025} & 0.647 $\pm$ 0.028 & 0.707 $\pm$ 0.021 & 0.751 $\pm$ 0.052 & 0.797 $\pm$ 0.035 & 0.596 $\pm$ 0.031 & 0.784 $\pm$ 0.019 & 7.468 $\pm$ 2.759 \\
\textbf{GeoMo-Net (Ours)} & \textbf{0.678 $\pm$ 0.011} & \textbf{0.739 $\pm$ 0.011} & \textbf{0.770 $\pm$ 0.036} & \textbf{0.826 $\pm$ 0.027} & \textbf{0.638 $\pm$ 0.028} & \textbf{0.810 $\pm$ 0.012} & \textbf{7.051 $\pm$ 2.962} \\
\hline
\end{tabular}
}
\end{table*}

\subsection{Data Acquisition and Pre-Processing}

We collected 225 paired short-axis LGE and multi-view cine MRI scans from 129 post-MI patients at the National University Hospital (NUH), Singapore.
All cardiac MRI examinations were performed on a 3.0 T scanner (Siemens Healthcare, Erlangen, Germany).
The cine MRI protocol included SAX balanced steady-state free precession (bSSFP) stacks and three standard long-axis (LAX) views (2-, 3-, and 4-chamber), comprising 8-17 slices and 25 temporal frames per sequence. 
Cine images were acquired with TE = 2.54 ms, temporal resolution = 48.24 ms, flip angle = $10^\circ$, slice thickness = 8.5 mm, slice spacing = 10.2 mm, field of view = $257\times320$ mm$^2$, and pixel bandwidth = 446 Hz.
LGE images were acquired with TR = 800 ms, TE = 1.56 ms, flip angle = $20^\circ$, slice thickness = 8.5 mm, slice spacing = 10.2 mm, field of view = $276\times340$ mm$^2$, and pixel bandwidth = 465 Hz.
The SAX LGE and cine MRI scans were spatially registered following the procedure described in Sec.~\ref{method:mapping}.  
At the patient level, the cohort had a mean age of $54.74 \pm 8.36$ years and a mean body mass index of $25.82 \pm 3.57$ kg/m$^2$, and a mean LV ejection fraction (LVEF) of $56.66 \pm 10.57\%$. 
Most patients underwent up to two imaging sessions separated by approximately 6-12 months, during which temporal remodeling of scar morphology may occur. 
Importantly, all scans from the same patient were assigned to the same fold to prevent data leakage between training, validation, and test sets. 
Specifically, we adopted a 12-fold cross-validation framework with an independent hold-out test set, where 180 scans from 98 patients were used for training and validation, and 45 scans from 31 patients were reserved for testing. 
To ensure fair comparison and minimize variability introduced by random partitioning, identical data splits were used across all experiments.

\begin{figure*}[t]
\centering
\includegraphics[width=0.92\linewidth]{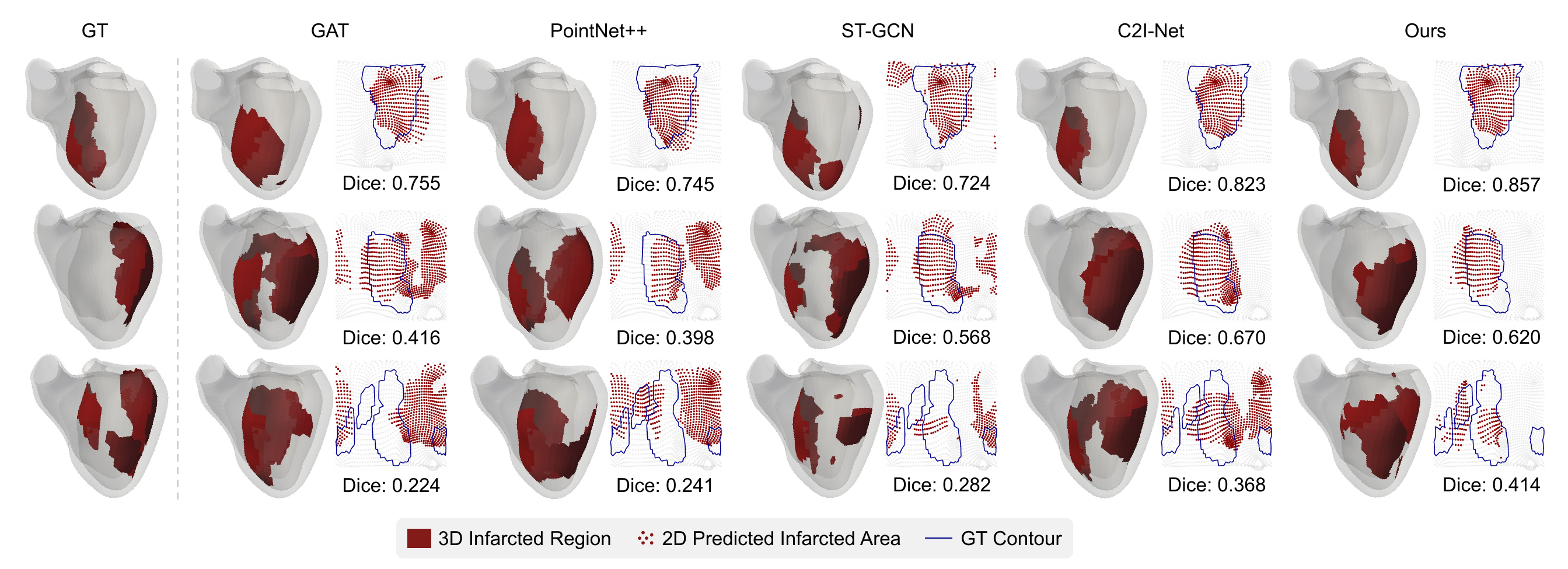}
\caption{Visualization of the scar reconstruction results using the five methods across three representative cases, corresponding to the top 10\% (best-performing), median, and bottom 10\% (most challenging) samples. Red regions on the biventricular geometry indicate scars. The 2D panels visualize the left ventricular (LV) endocardial point cloud in the relative universal ventricular coordinate (UVC) space for intuitive comparison, where red dots denote predictions and blue contours indicate the ground truth (GT). }
\label{fig:result:comparison}
\end{figure*}
\subsection{Implementation Details}

All experiments were implemented in PyTorch and conducted on a workstation equipped with an AMD EPYC 7K62 CPU and an NVIDIA GeForce RTX 4090 GPU.
We used the AdamW optimizer to update the network parameters. 
The initial learning rate is set to $1 \times 10^{-3}$ and the weight decay is set to $1 \times 10^{-4}$.
Within the 12-fold cross-validation framework, each fold was trained for a maximum of 200 epochs using a batch size of 16.
The balancing parameters in Sec.~\ref{method:recon} are set as follows:
$\lambda_{\text{focal}} = 1.8$ and $\lambda_{\text{coarse}} = 4$.
Furthermore, based on the results from the validation set, the model training adopts an early stopping strategy with a patience of 20 rounds.
At the same time, the loss and Dice score of the validation set are monitored to converge to the best model.
The training time for the 12-fold model is approximately 2.5 hours, while the average inference time for each case is approximately 14 ms.

\begin{table*}[htbp]
\centering
\caption{Quantitative results of ablation studies evaluating different components of the proposed method for the 3D infarct reconstruction.}
\label{exp:tb:ablation_study}
{\scriptsize
\begin{tabular}{l| c c c c c c c}
\hline
Method & Dice $\uparrow$ & G-Dice $\uparrow$ & Sensitivity $\uparrow$ & Specificity $\uparrow$ & Precision $\uparrow$ & Accuracy $\uparrow$ & ASSD (mm) $\downarrow$\\
\hline
\textbf{w/o} Geo    & 0.601 $\pm$ 0.024 & 0.667 $\pm$ 0.029 & 0.732 $\pm$ 0.033 & 0.753 $\pm$ 0.055 & 0.545 $\pm$ 0.047 & 0.747 $\pm$ 0.033 & 8.004 $\pm$ 2.797 \\
\textbf{w/o} Mo     & 0.631 $\pm$ 0.037 & 0.698 $\pm$ 0.038 & 0.737 $\pm$ 0.049 & 0.792 $\pm$ 0.062 & 0.591 $\pm$ 0.059 & 0.777 $\pm$ 0.039 & 7.952 $\pm$ 2.987 \\
\textbf{w/o} AHA 17 & 0.663 $\pm$ 0.032 & 0.724 $\pm$ 0.034 & 0.752 $\pm$ 0.040 & 0.816 $\pm$ 0.057 & 0.626 $\pm$ 0.059 & 0.798 $\pm$ 0.035 & 7.323 $\pm$ 2.626 \\
\textbf{GeoMo-Net} & \textbf{0.678 $\pm$ 0.011} & \textbf{0.739 $\pm$ 0.011} & \textbf{0.770 $\pm$ 0.036} & \textbf{0.826 $\pm$ 0.027} & \textbf{0.638 $\pm$ 0.028} & \textbf{0.810 $\pm$ 0.012} & \textbf{7.051 $\pm$ 2.962} \\
\hline
\end{tabular}
}
\end{table*}

\subsection{Gold Standard and Evaluation}

Infarcted regions were manually delineated by experienced experts on the LGE MRI scans. 
These pixel-wise annotations were subsequently mapped onto the reconstructed 3D biventricular mesh to establish node-level ground truth labels. 
The node-wise labels were further aggregated according to the AHA 17-segment model to obtain reference segment-level scar burdens for coarse-scale supervision and regional evaluation.
At the node level, 3D infarct reconstruction performance was quantitatively evaluated using the Dice coefficient, generalized Dice (G-Dice) \citep{conf/MICCAI-DLMIA/sudre2017}, sensitivity, specificity, precision, accuracy, and average symmetric surface distance (ASSD).
Specifically, G-Dice incorporates all classes with class-specific weights, where rarer classes are assigned larger weights to mitigate the effect of class imbalance. 
ASSD computes the shortest Euclidean distance from each sampled point on the GT boundary to the predicted boundary, and vice versa, ultimately averaging all the shortest distances from both directions.
These metrics assess both spatial overlap and node-wise classification performance between the predicted scar regions and LGE-derived ground truth.
Furthermore, we assessed the consistency of scar localization and burden estimation. 
Specifically, within the UVC system, the Mean Absolute Error (MAE) was employed to quantify deviations in scar centroid location and scar burden volume relative to the ground truth. 
The Pearson correlation coefficient (PCC) was additionally computed to evaluate the overall agreement between predicted and reference scar distributions.

\subsection{Comparison Study} \label{exp:result:comparison}

To comprehensively evaluate the effectiveness of GeoMo-Net for 3D scar reconstruction, we compared the proposed framework against four representative baseline methods including graph attention network (GAT) \citep{conf/ICLR/velickovic2018}, PointNet++ \citep{conf/NIPS/qi2017}, ST-GCN \citep{conf/AAAI/yan2018}, C2I-Net \citep{conf/DT4H/lyu2025}, as presented in Table~\ref{exp:tb:comparison}.
Here, GAT was used as a graph-based spatial baseline, whereas PointNet++ served as a topology-free point-cloud baseline.
ST-GCN was included to evaluate conventional spatio-temporal graph modeling. 
C2I-Net served as a strong task-specific baseline for evaluating the improvements introduced by the proposed geometry-motion decoupling strategy and enhanced spatio-temporal modeling framework. 
Overall, GeoMo-Net consistently achieved the best performance across all evaluation metrics, demonstrating superior capability in reconstructing node-level infarct distributions from cine-derived cardiac motion. 
Compared with the strongest baseline (C2I-Net), the proposed method improved the average Dice score from 0.647 to 0.678 and the generalized Dice score from 0.707 to 0.739, indicating substantially improved overlap between the predicted infarct regions and ground truth. 
GeoMo-Net also achieved the highest sensitivity (0.770 $\pm$ 0.036) and precision (0.638 $\pm$ 0.028), suggesting improved identification of infarcted regions while simultaneously reducing false positive predictions.

The qualitative comparisons in Fig.~\ref{fig:result:comparison} further demonstrate the advantages of the proposed framework across representative cases ranked by case-wise Dice score, including top 10\%, median, and bottom 10\% cases. 
In the top-performing cases, the predicted scar regions from GeoMo-Net almost completely overlap with the ground truth, accurately preserving the scar location, shape, and extent. 
The corresponding UVC visualizations further show strong alignment between the predicted distributions and the reference contours, indicating that GeoMo-Net can capture localized scar patterns in the anatomical coordinate domain. 
For the median cases, GeoMo-Net successfully reconstructs the overall scar topology and macroscopic distribution, with minor deviations along local boundaries. 
In this tier, C2I-Net also demonstrates strong qualitative performance and achieves highly competitive reconstructions, suggesting that both frameworks can effectively capture dominant infarct patterns in relatively typical cases.
The advantages of GeoMo-Net become more evident in the challenging bottom 10\% cases. 
Although some omissions remain around highly irregular boundary regions, the proposed framework can still preserve the overall scar location and major topological structure. 
In contrast, other baseline methods, particularly GAT and PointNet++, exhibit more severe regional misclassifications, fragmented predictions, and topological inconsistencies. 
ST-GCN improves spatial continuity to some extent through temporal modeling, but still struggles with sparse and heterogeneous scar distributions. 
Visual inspection suggests that these difficult cases are typically associated with extreme scar sizes, highly complex scar topologies, or marginal anatomical locations near the basal or apical regions. 
Nevertheless, GeoMo-Net demonstrates substantially improved robustness in these scenarios, highlighting the benefit of jointly modeling geometric topology and temporal cardiac motion for anatomically consistent myocardial scar reconstruction.

\begin{figure*}[t]
\centering
\includegraphics[width=0.88\textwidth]{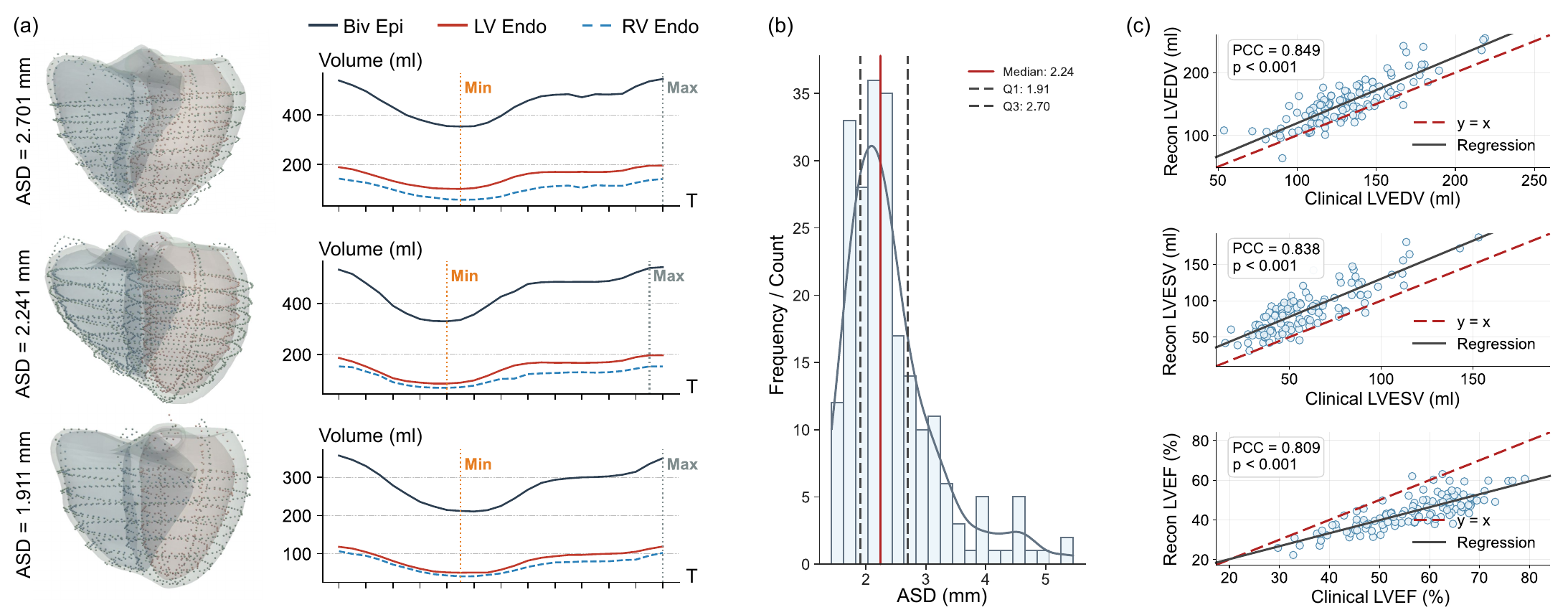}\\[-2ex]
\caption{Evaluation of 4D biventricular mesh reconstruction quality. (a) Visual comparison of three representative cases corresponding to lower quartile, median, and upper quartile of the average surface distance (ASD) distribution. The right panels show the corresponding temporal volume variations of the biventricular epicardium (Biv Epi), LV endocardium (Endo), and right ventricular (RV) endocardium throughout the cardiac cycle, with the minimum and maximum volumes indicated by dashed lines. (b) Distribution of ASD across all cases, with the representative cases highlighted. (c) Correlation analysis between mesh-derived and clinical measurements of LV end-diastolic volume (LVEDV), LV end-systolic volume (LVESV), and LV ejection fraction (LVEF). PCC: Pearson correlation coefficients. Recon: Reconstruction.}
\label{fig:result:4dheart_recon}
\end{figure*}

\subsection{Ablation Study} \label{exp:result:ablation}

To systematically evaluate the contribution of each core component in GeoMo-Net, we conducted a series of ablation studies, as presented in Table~\ref{exp:tb:ablation_study}. 
Overall, the complete GeoMo-Net achieved the best performance across all metrics, demonstrating the effectiveness of jointly modeling geometric topology, temporal motion dynamics, and anatomical prior supervision for myocardial scar reconstruction.
Among all ablation settings, removing the geometry branch resulted in the most substantial performance degradation, with Dice decreasing from 0.678 to 0.601 and accuracy dropping from 0.810 to 0.747. 
This observation highlights the fundamental importance of preserving myocardial surface geometry and topology for accurate scar localization. 
Similarly, excluding the motion branch led to noticeable reductions in Dice (0.631) and Sensitivity (0.737), demonstrating that dynamic cardiac deformation patterns provide complementary pathological information beyond geometry features alone. 
Furthermore, removing the AHA 17-segment anatomical prior also degraded performance, particularly in Sensitivity and Dice-related metrics.
This indicates that coarse-grained regional supervision provides additional anatomical regularization for node-level predictions and improves the spatial consistency of reconstructed scar distributions.
Although the quantitative improvements introduced by the AHA prior are relatively moderate compared with geometry and motion features, this component remains important for clinical interpretability. 
In routine clinical practice, the AHA 17-segment model serves as the standardized representation for myocardial scar assessment and reporting, providing an interpretable and clinically meaningful description of infarct distribution \citep{journal/JCMR/schulz2020}. 
Incorporating this anatomical prior therefore not only improves model robustness, but also enhances the clinical interpretability and compatibility of the reconstructed scar maps with existing clinical workflows. 

\subsection{Results}

\subsubsection{Accuracy of 4D Biventricular Reconstruction} \label{exp:result:4d_heart}

We quantitatively evaluated the spatial fidelity of the reconstructed 4D biventricular meshes against manual segmentation contours across all subjects. 
The proposed framework achieved an average surface distance (ASD) of $2.43 \pm 0.75$ mm, indicating high geometric agreement between the reconstructed meshes and the reference annotations. 
Fig.~\ref{fig:result:4dheart_recon} (a) presents three representative cases corresponding to the lower quartile, median, and upper quartile of the ASD distribution. 
Overall, the reconstructed meshes preserve smooth and anatomically plausible ventricular geometries throughout the cardiac cycle. 
The reconstructed surfaces also align well with the sparse endocardial annotations, demonstrating strong morphological consistency.
The corresponding volume-time curves further show that the reconstructed 4D meshes can continuously capture systolic contraction and diastolic relaxation dynamics. 
Physiological temporal variations of the biventricular epicardial, LV endocardial, and RV endocardial volumes are consistently preserved across the cardiac cycle. 
In addition, Fig.~\ref{fig:result:4dheart_recon} (b) summarizes the ASD distribution across all subjects, where most cases are concentrated around low reconstruction errors, suggesting stable and robust reconstruction performance across the cohort.

To further assess physiological reliability, we compared mesh-derived cardiac functional metrics against standard clinical MRI measurements, including LV end-diastolic volume (LVEDV), LV end-systolic volume (LVESV), and LVEF, as shown in Fig.~\ref{fig:result:4dheart_recon} (c). 
Strong positive correlations were observed between the reconstructed and clinical measurements (LVEDV: $r = 0.849$; LVESV: $r = 0.838$; LVEF: $r = 0.809$; all $p < 0.001$). 
These results demonstrate that the proposed framework can reliably capture subject-specific cardiac function and dynamic ventricular geometry. 
Although the reconstructed LVEDV and LVESV showed a systematic overestimation relative to clinical measurements, leading to a corresponding underestimation of LVEF, the overall physiological trends and inter-subject functional relationships were well preserved. 
Consistent with previous studies \citep{bellenger2000comparison}, these discrepancies are likely caused by methodological differences in volumetric quantification, particularly ambiguities in basal plane definition and variations in endocardial boundary delineation during automated mesh reconstruction. 
Nevertheless, the strong correlations and stable temporal dynamics indicate that the reconstructed 4D meshes provide physiologically meaningful representations of cardiac motion and function.

\begin{figure*}[t]
\centering
\includegraphics[width=1\textwidth]{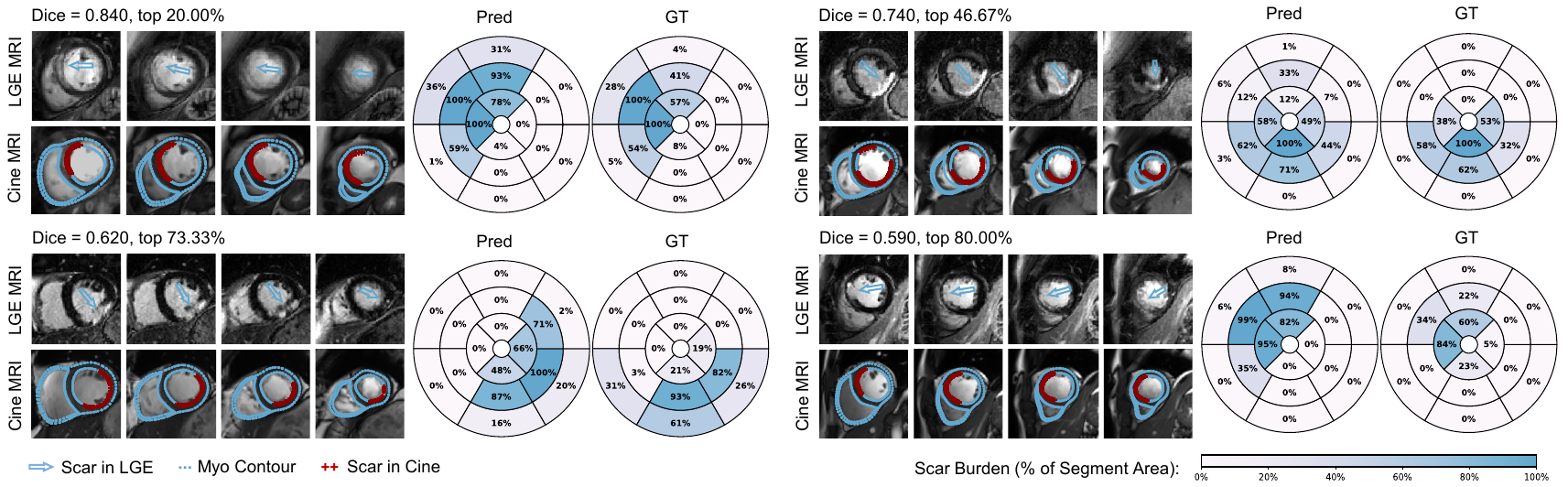}\\[-2ex]
\caption{2D scar prediction results in four representative cases. For each case, the upper-left panels show the LGE MRI slices with enhanced scar regions indicated by arrows, while the lower-left panels present the corresponding cine MRI overlaid with myocardial contours and scar predictions back-projected from the 3D reconstructed infarct. The bull's-eye plots compare the predicted segment-level scar burden with the corresponding GT based on the AHA segment model.}
\label{fig:result:scar_2dvisual}
\end{figure*}

\begin{figure*}[t]
\centering
\includegraphics[width=0.88\textwidth]{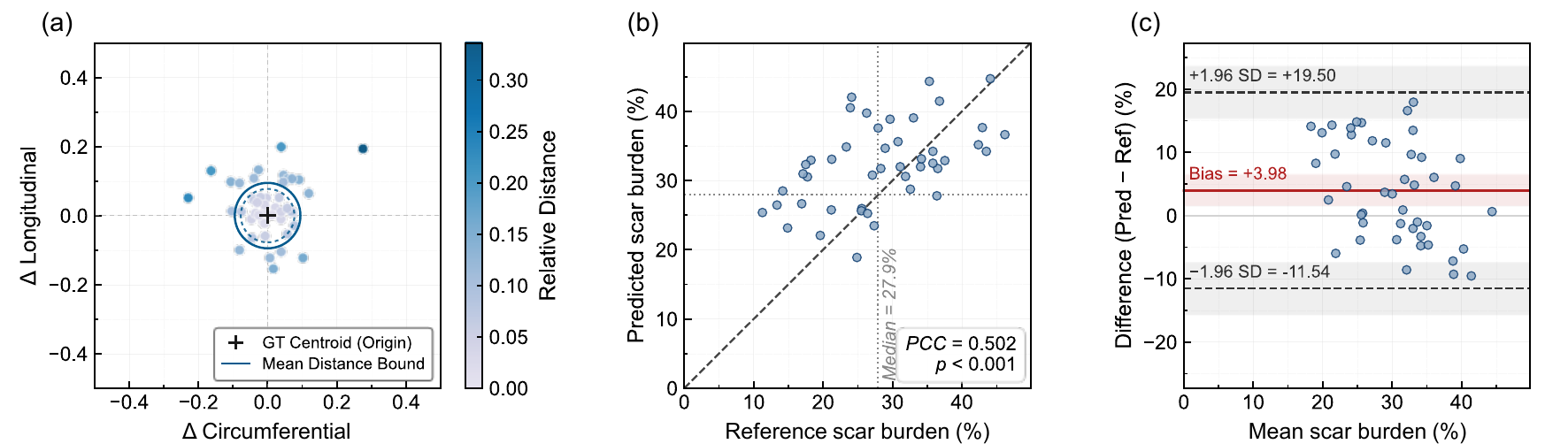}\\[-2ex]
\caption{Evaluation of scar localization and burden estimation. (a) Spatial displacement of scar centroids between the predicted and ground-truth infarct regions in the circumferential-longitudinal coordinate space. Each point represents one case, and the dashed circle indicates the mean normalized Euclidean localization error. (b) Correlation analysis of scar burden between the predictions and the ground truth. (c) Corresponding Bland-Altman analysis of scar burden estimation.}
\label{fig:result:scar_size}
\end{figure*}

\subsubsection{Accuracy of 3D Infarct Reconstruction} \label{exp:result:3d_infarct}

To provide a reference for the quantitative evaluation, we assessed inter-observer variation using three independent manual scar delineations. 
Fifteen cases were randomly selected and annotated separately by three experts, and the resulting scar labels were projected onto the surface mesh for comparison. 
The Dice coefficients obtained from the three observers were $0.757 \pm 0.110$, $0.756 \pm 0.090$, and $0.744 \pm 0.058$, respectively, indicating comparable annotation consistency across observers and reflecting the inherent ambiguity of manual scar delineation.
Table~\ref{exp:tb:comparison} summarizes the quantitative performance of GeoMo-Net for 3D scar reconstruction. 
The proposed framework achieved a Dice of 0.678 $\pm$ 0.011, which is comparable to the inter-observer variation. 

To further evaluate regional modeling consistency, we analyzed scar distributions using the standardized AHA segment model.
In visualization, the apical cap was excluded, resulting in a 16-segment representation \citep{journal/EHRCI/lang2015}, as illustrated in Fig.~\ref{fig:result:scar_2dvisual}.
Four representative cases with varying scar locations, sizes, and reconstruction quality were selected for visualization. 
In each bull's-eye map, the scar burden within each segment is color-coded according to the percentage of infarct involvement. 
Despite potential inaccuracies introduced by the registration between LGE and cine MRI, the predicted scar distributions demonstrate strong regional correspondence with the ground-truth maps. 
In particular, the major infarct territories, spatial extent, and segment-level scar patterns remain highly consistent across most cases, indicating that the proposed framework can reliably recover clinically meaningful regional scar distributions from cine-derived motion information. 
Quantitatively, the predicted and ground-truth segment-level scar burdens exhibited a strong linear correlation (PCC = 0.769, $p < 0.001$), demonstrating substantial structural consistency at the regional level. 
Using the global mean scar burden (28.92\%) as the threshold for segment-level binary classification, the proposed method achieved an overall accuracy of 0.822, with a sensitivity of 0.767 and a specificity of 0.856, indicating a balanced performance between false negatives and false positives. 
In addition, the F1-score reached 0.767 and the Cohen's Kappa coefficient was 0.623, further confirming substantial agreement between the predicted regional scar distributions and the clinical reference standard.

Furthermore, we evaluated the spatial localization accuracy and scar burden estimation of the reconstructed infarct regions, as illustrated in Fig.~\ref{fig:result:scar_size}. 
For spatial localization analysis, the centroid displacement between the predicted scars and the ground truth was quantified within the circumferential-longitudinal coordinate space, as shown in Fig.~\ref{fig:result:scar_size} (a). 
The proposed framework achieved mean relative localization errors of 5.81\% and 6.43\% along the circumferential and longitudinal directions, respectively, corresponding to an overall relative error of 9.48\%. 
Most predicted scar centroids were concentrated near the ground-truth origin, indicating strong spatial consistency between the reconstructed and reference infarct distributions. 
These results suggest that GeoMo-Net can provide anatomically consistent scar localization, which is important for downstream patient-specific EP simulation.
We further assessed scar burden estimation using correlation and Bland-Altman analyses, as shown in Fig.~\ref{fig:result:scar_size} (b)-(c). 
GeoMo-Net achieved a scar burden prediction MAE of $7.24\% \pm 5.03\%$, with a moderate but significant correlation with the reference standard (PCC = 0.502, $p < 0.001$). 
The Bland-Altman analysis demonstrated an average bias of $3.98\%$, with limits of agreement ranging from $-11.54\%$ to $19.50\%$.
Overall, the proposed framework effectively captured the global trend of infarct burden variation across patients. 
In particular, the model demonstrated relatively strong sensitivity for identifying larger infarct regions, while prediction variability remains higher in cases with relatively small scar burdens. 
Despite these deviations, the predicted scar burden distributions remained generally consistent with the reference measurements in this cohort.

\begin{figure}[t]
\centering
\includegraphics[width=0.95\linewidth]{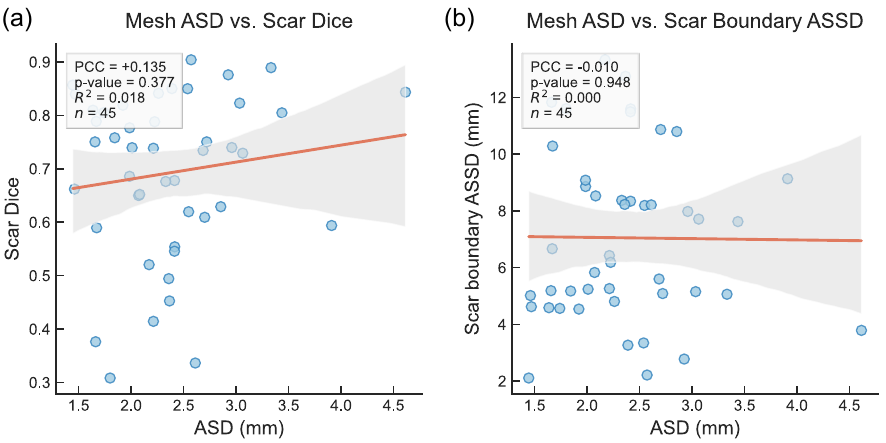}\\[-2ex]
\caption{Correlation analysis between 4D biventricular reconstruction error in terms of ASD and 3D scar reconstruction accuracy in terms of Dice and ASSD. Scatter distribution and linear fitting of the raw values.}
\label{fig:result:correlation}
\end{figure}

\begin{figure*}[t]
\centering
\includegraphics[width=1\textwidth]{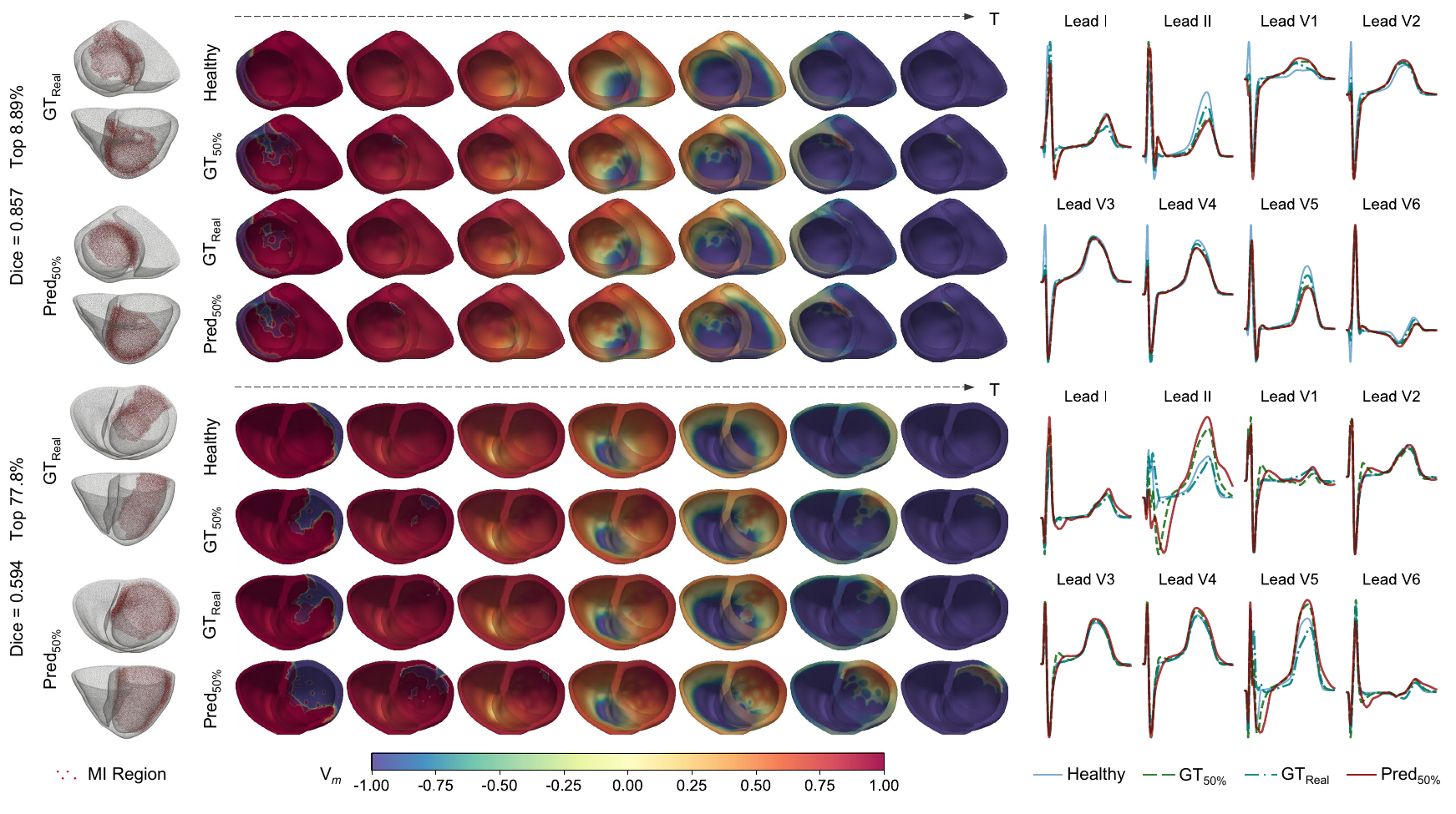}\\[-2ex]
\caption{Patient-specific electrophysiological simulation under different scar configurations. (a) Patient-specific biventricular mesh shown from two views, with myocardial scar overlaid on the ventricular geometry.
(b) Transmembrane potential maps ($V_m$) illustrating scar-induced alterations in ventricular activation and conduction patterns across four different structural configurations. 
(c) Simulated 12-lead ECG signals under different settings. Here, \textit{Healthy} assumed no scar on the biventricular geometry, \textit{GT$_{50\%}$} used LGE-derived scar position with 50\% transmurality, \textit{GT$_{Real}$} used LGE-derived scars with patient-specific transmurality, and \textit{Pred$_{50\%}$} applied GeoMo-Net predicted scars with 50\% transmurality.}
\label{fig:result:simulation}
\end{figure*}

\subsubsection{Correlation between 4D Biventricular and 3D Infarct Reconstruction} \label{exp:result:correlation}

To investigate the relationship between 4D biventricular reconstruction quality and 3D infarct reconstruction performance, we analyzed the correlation between the ASD of the reconstructed cardiac meshes and two infarct evaluation metrics: the Dice score (region overlap) and the boundary ASSD (surface distance), as illustrated in Fig.~\ref{fig:result:correlation}.  
Fig.~\ref{fig:result:correlation} (a) and (b) present the scatter distributions and linear fittings for mesh ASD vs. scar Dice and mesh ASD vs. scar boundary ASSD, respectively. 
Overall, the results demonstrate a near-zero association between the geometric mesh reconstruction error and infarct prediction performance.
Specifically, the correlation analysis for scar Dice yielded a Pearson $R^2$ value of 0.018 with a non-significant $p$-value of 0.377, indicating that the variability in infarct reconstruction accuracy cannot be effectively explained by the mesh reconstruction error alone.
More notably, mesh ASD showed no clear linear association with scar boundary ASSD ($R^2 < 0.001$, $p = 0.948$).
These findings strongly suggest that moderate variations in 4D mesh reconstruction quality do not lead to degradation in infarct reconstruction performance.
One possible explanation is that the proposed framework primarily relies on the global spatio-temporal deformation patterns and intrinsic cardiac motion representations, which remain relatively robust even in the presence of small geometric reconstruction inaccuracies. 
In addition, myocardial scar reconstruction is more strongly influenced by pathological motion abnormalities and regional deformation characteristics than by minor surface reconstruction errors. 
Therefore, although accurate 4D biventricular reconstruction provides the anatomical foundation for subsequent analysis, the final infarct prediction performance appears to depend more critically on the learned motion-pathology representations rather than purely geometric reconstruction fidelity.

\subsection{In-Silico Electrophysiological Simulation Evaluation} \label{exp:result:simulation}

To assess the downstream functional utility of GeoMo-Net within the CDT framework, we performed an in-silico EP simulation study.
We employed the pseudo-domain reaction-eikonal model \citep{journal/MedIA/camps2025}, with the Mitchell-Schaeffer model \citep{journal/BMB/Mitchell2003} as the reaction operator.
Four structural configurations were compared: \textit{Healthy}, \textit{GT$_{50\%}$}, \textit{GT$_{Real}$}, and \textit{Pred$_{50\%}$}.
Note that 50\% subendocardial scar is typically used as a practical approximation of moderate MI transmurality when patient-specific transmurality is unavailable \citep{journal/Circ/reimer1977,journal/IJNMBE/leong2017}.
As shown in Fig.~\ref{fig:result:simulation}, two representative cases with different scar burdens and prediction accuracy were evaluated. 
The predicted scars were overlaid on the patient-specific biventricular meshes and compared with the LGE-derived ground truth from two anatomical views. 
Across both cases, the predicted scar distributions produced transmembrane potential ($V_m$) propagation patterns that closely followed the LGE-based simulations, including scar-induced conduction delay and local conduction block. 
In contrast, the healthy baseline exhibited smoother and more homogeneous activation patterns without regional propagation disturbance.
The simulated ECGs further demonstrate the functional relevance of the predicted scar substrate. 
Compared with the healthy baseline, GeoMo-Net-based simulations reproduced lead-specific morphological abnormalities observed in the ground-truth scar models, including altered QRS morphology and ST-segment deviations. 
Although the predicted model uses a subendocardial infarction pattern rather than patient-specific transmurality, its ECG and $V_m$ patterns remain substantially closer to the $GT_{real}$ simulation than to the healthy baseline.
This suggests that, even with simplified transmural properties, the predicted 3D scar topology provides meaningful EP information. 
Moreover, the GT$_{50\%}$ condition generally lies between the healthy and GT$_{real}$ responses, indicating that residual discrepancies are partly attributable to the simplified subendocardial transmural assumption rather than scar localization alone. 
These results suggest that GeoMo-Net recovers a spatially resolved EP substrate capable of inducing anatomically consistent conduction abnormalities and lead-specific ECG signatures, even under approximated transmural properties.

\subsection{External Validation on the Public Dataset} \label{exp:result:external_validation}

To evaluate the generalizability of GeoMo-Net, we performed external validation on the public CMR-MULTI dataset \citep{arxiv/qu2026}, which provides multi-view cine MRI and LGE MRI. 
Among 105 released training cases, 30 contained scar labels, and 10 remained after excluding cases with unsuccessful biventricular mesh reconstruction due to absence of the standard 3-chamber cine MRI view. 
These cases were processed using the same reconstruction and inference pipeline as the internal cohort, and all 12 internally trained fold-specific models were directly evaluated without fine-tuning. 
GeoMo-Net achieved a Dice score of $0.476 \pm 0.148$, while the predicted scars generally preserved the dominant anatomical orientation and regional distribution of the ground truth. 
Although lower than the internal test performance, this gap is likely due to compounded domain shifts, including imaging protocols and a markedly different infarction distribution. 
Specifically, the valid external cases showed lower mean LVEF than the internal cohort ($43.89\%$ vs. $56.38\%$) and substantially higher scar burden ($60.85\%$ vs. $28.30\%$), placing several cases beyond the predictive range covered by our training data. 
Despite these challenges, GeoMo-Net retained meaningful scar localization capability under external domain shift, providing preliminary evidence of cross-domain generalizability while highlighting the need for larger multi-center training cohorts, more robust reconstruction under incomplete cine views, and improved coverage of high-burden scar phenotypes.

\section{Discussion and Conclusion}

In this study, we presented GeoMo-Net, a fully automatic framework for reconstructing personalized, scar-embedded 3D biventricular models directly from routine multi-view cine MRI for cardiac digital twin modeling. The proposed method explicitly decouples geometry-aware and motion-aware representations on topology-consistent 4D meshes, enabling direct mapping from cine-derived myocardial deformation to 3D infarct geometry. The proposed model was evaluated on 129 subjects (225 scans) from NUH Singapore.
It consistently outperformed graph-based, point-cloud-based, and spatio-temporal baselines, achieving the best Dice score of $0.678 \pm 0.011$, generalized Dice of $0.739 \pm 0.011$, and accuracy of $0.810 \pm 0.012$ (Sec. \ref{exp:result:comparison}). The ablation study further confirmed the importance of both geometric topology and motion dynamics, while the AHA-17 segment prior improved regional consistency and clinical interpretability (Sec. \ref{exp:result:ablation}). Beyond node-level overlap, GeoMo-Net produced anatomically meaningful scar reconstructions, with strong AHA segment-level agreement, accurate scar centroid localization, and scar burden estimates that followed the reference trend, as demonstrated in Sec. \ref{exp:result:3d_infarct}. Downstream EP simulations further showed that the predicted scars induced conduction delays and lead-specific ECG abnormalities consistent with LGE-derived ground truth, demonstrating the functional relevance of the reconstructed infarct geometry (Sec. \ref{exp:result:simulation}). External validation on the newly public CMR-MULTI dataset also provided preliminary evidence that the model retains meaningful localization capability under domain shift, although with reduced accuracy in more challenging high-burden cases (Sec. \ref{exp:result:external_validation}).

Despite these advances, this study still has several limitations. 
First, the current framework predicts the spatial distribution of infarcted regions but does not recover patient-specific scar transmurality, which is a highly challenging task from cine MRI alone because transmural information is only indirectly reflected through myocardial motion. In the simulation study, we therefore used a simplified 50\% subendocardial scar assumption when patient-specific transmurality was unavailable. Although this approximation still produced meaningful EP patterns closer to the ground truth than the healthy baseline, it may not fully capture the heterogeneous intramural structure of real infarcts. 
Second, the ground-truth scar labels were derived from sparse 2D LGE MRI slices and required registration and projection onto the cine-derived mesh, which may introduce uncertainty in boundary regions. 
Third, external validation was limited by domain shift, incomplete cine views, and a small number of usable public cases, indicating that robustness across centers, scanners, and high-scar-burden phenotypes requires further validation.

Future work will focus on extending GeoMo-Net toward more comprehensive and clinically deployable MI digital twins. A key direction is to estimate scar transmurality and heterogeneous border-zone properties, potentially by combining cine MRI, ECG, and clinical biomarkers. We will also expand training and validation to larger multi-center cohorts with broader disease severity, acquisition protocols, and scar phenotypes, while improving mesh reconstruction robustness under missing or low-quality cine views. Finally, we plan to integrate the reconstructed scar-embedded models with patient-specific ECG calibration and faster surrogate EP solvers, enabling more efficient risk stratification, virtual pacing, and treatment planning for post-MI patients. 
In conclusion, this work demonstrates the feasibility of directly reconstructing simulation-ready 3D infarct geometry from contrast-free cine MRI, providing an important step toward scalable CDTs for post-MI assessment and personalized EP simulation.

\bibliographystyle{model2-names}
\biboptions{authoryear}
\bibliography{A_refs}

\end{document}